\documentclass{article} 
\usepackage{iclr2027_conference,times}

\usepackage{amsmath,amsfonts,bm}

\def\eqref#1{equation~\ref{#1}}

\def\1{\bm{1}}

\DeclareMathAlphabet{\mathsfit}{\encodingdefault}{\sfdefault}{m}{sl}
\SetMathAlphabet{\mathsfit}{bold}{\encodingdefault}{\sfdefault}{bx}{n}

\usepackage{hyperref}
\usepackage{url}
\usepackage{graphicx} 
\usepackage{subcaption} 
\usepackage[utf8]{inputenc}
\usepackage{newunicodechar}
\newunicodechar{−}{\textminus}

\title{Modeling bilingual naming performance through laterally-connected SOMs}

\author{Divya Prabhakaran, Uli Grasemann, Risto Mikkulainen  \\
Department of Computer Science\\
The University of Texas at Austin\\
\texttt{\{divyapr,uli,risto\}@cs.utexas.edu} \\
\And
Erie Shivers, Marissa Russell-Meill \& Swathi Kiran \\
Center for Brain Recovery \\
Boston University \\
\texttt{\{erinhana,mruss,kirans\}@bu.edu} \\
}

\iclrfinalcopy 
\begin{document}
\maketitle

\begin{abstract}
Computational models have explanatory power to understand cognition in individuals, but there is usually not enough training data to build such models with traditional machine learning. However, alternative architectures that model neural and cognitive processes in more detail can be more easily fit to individuals and can lead to mechanistic insights into human behavior. BiLex is such a computational model of the bilingual lexicon, containing a shared semantic system and phonetic language systems represented through self-organizing maps. In this study, BiLex was extended further into biological realism with excitatory and inhibitory lateral connections. This extension makes it possible to account for response times as well as common errors such as those based on semantic concept familiarity, typicality, and specificity. The Boston Naming Test was simulated in Spanish and English, and model performance approximated naming accuracy and error types in Spanish-English bilinguals with varying primary and secondary language histories. The results indicated that the approach captures semantic-level processes in bilingual lexical access, thus modeling the semantic cognition of individual subjects. In the future, such models may serve as a foundation for modeling impairments and treatment options.
\end{abstract}

\section{Introduction}
\begin{figure}
\centering
\includegraphics[width=1.0\linewidth]{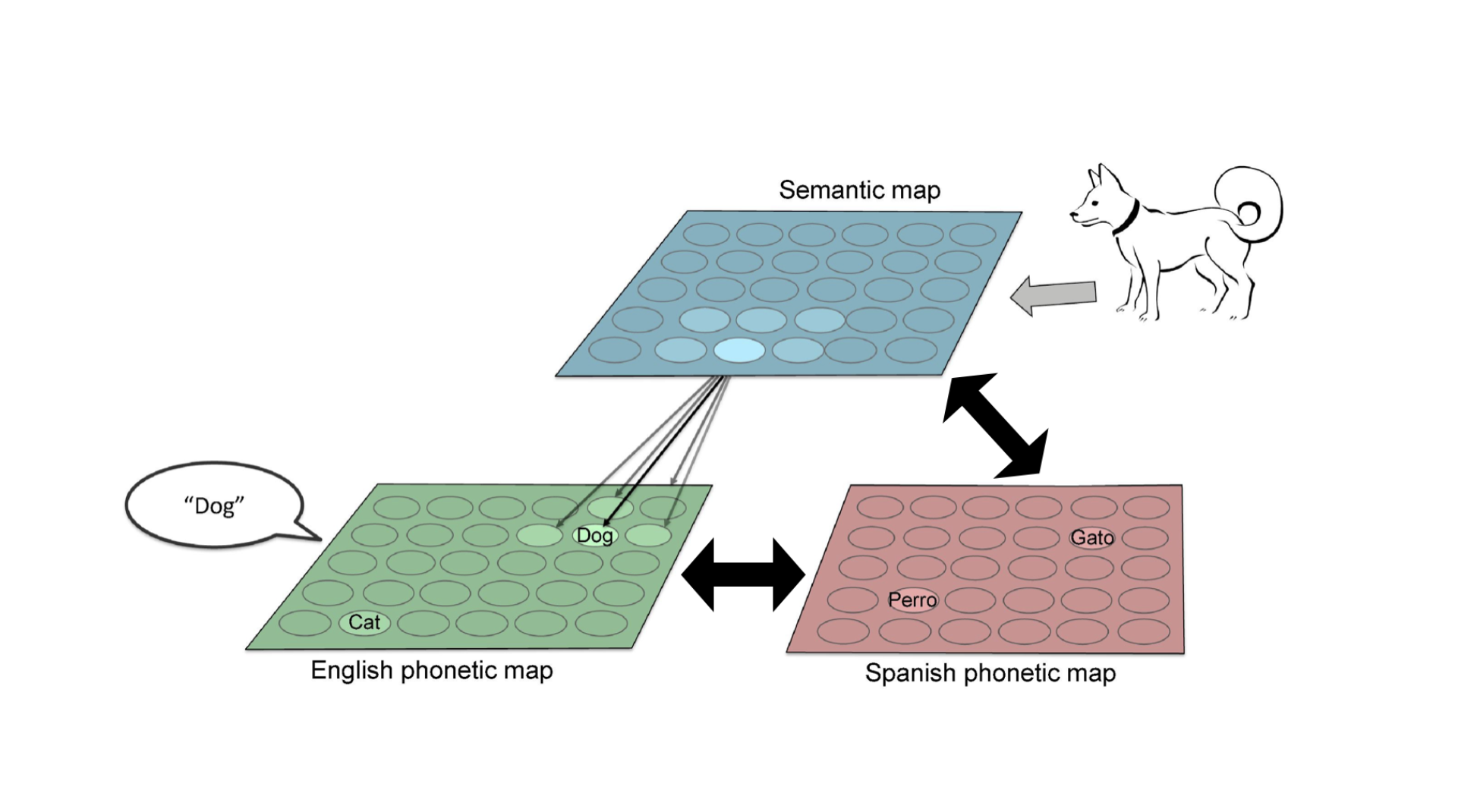}\\[-8ex]
\caption{Laterally-connected BiLex model. The model simulated naming by receiving input to the semantic map, propagating activation through lateral connections and associative connections across all maps, and producing a response from the target language map. This architecture is motivated by the lexical-semantic structures in the brain, and can therefore serve as a foundation for modeling individualized performance, its breakdown in e.g.\ stroke and dementia, and identifying effective treatments \citep{Kiran2026}. Figure adapted from \citep{Penaloza2019}}
\label{fig:bilexmodel}
\end{figure}

The BiLex model is built on self-organizing maps (SOMs), which use competitive learning to capture the underlying structure of high-dimensional input data in a low-dimensional, 2D map representation \citep{Kohonen1995}. The bilingual language system is represented by three SOMs: two phonetic (lexical) maps representing Spanish and English words, and a semantic map representing shared word meanings \citep[Figure~\ref{fig:bilexmodel};][]{Penaloza2019}. Associative connections between these maps implement functions such as naming, comprehension, and translation.

This work extends the BiLex model with lateral connections. Whereas the response in the traditional SOM is determined instantaneously as the map unit closest to the input, the lateral connections in this extended model settle the response over time through short-range excitation and long-range inhibition. Inhibitory connections are uniformly distributed across the map neurons, and excitatory connections are focused on a Gaussian neighborhood with a standard deviation $\sigma$ that reduces over time. With this model, it is possible to account for response times and produce a variety of error behaviors, which can then be matched with those produced by human subjects.

In a series of simulations, the extended BiLex model was trained to model subjects whose L1 was either English or Spanish, and whose L2 acquisition took place early or late.  Further, word familiarity was modeled with word frequency, typicality with map organization, and specificity with superordinate-subordinate associative connections. The resulting model accurately replicated the subject's naming performance and types of errors. It can therefore serve as a foundation for building individualized models of bilingual lexical-semantic processing, potentially leading to more accurate treatment recommendations for impairments such as stroke and dementia than prior work \citep{Kiran2026}.

\section{Methods}
The BiLex model was expanded to account for semantic effects such as familiarity, typicality, and specificity. All maps and associative connections were trained together in joint models, with different primary (L1) and secondary (L2) bilingual language usage conditions.

\subsection{Semantic and Phonetic Data.}
The word corpus (792 words) consisted of 128 superordinate words (e.g.\ categories such as "animal") and 664 subordinate words (e.g.\ "horse"). Phonetic feature representations were obtained through the same method as with previous BiLex models: phonetic representations were based on feature-based encodings of their IPA transcriptions \citep{Penaloza2019, Fidelman2022, prabhakaran:cogsci25}. Encodings include phonemes, vowels, and consonant properties. GPT-4o \citep{OpenAI2024} was used to obtain and define semantic feature values for all subordinate and superordinate words. A total of 664 subordinate words $\times$ 682 features were collected. Responses of "1" for yes and "0" for no formed the semantic feature vector for each subordinate word. GPT-based semantics were compared to MTurk data for earlier BiLex models, where cognitively healthy adults were asked to give yes/no responses if a word had a semantic feature. Validation analysis found $>90\%$ agreement between MTurk and GPT-based semantics. The resulting semantic and phonetic feature vectors were used in map training, with both superordinate and subordinate words represented in the lexical maps and only subordinate words represented in the semantic map. 

Word frequency was used as a proxy for familiarity in the semantic and lexical maps. Frequencies were primarily obtained from the ClearPond database \citep{Marian2012}. Corpus of Contemporary American English (COCA) \citep{Davies2008} was used for English words missing in ClearPond. Missing Spanish word frequencies were filled in using those of their English counterparts, which were found to be sufficiently predictive in the relevant frequency range ($R^2=0.584; p\ll 0.001$). Typicality of a concept can be defined as how well a concept represents its semantic category \citep{typicalityPatterson}. Typicality values were obtained by prompting GPT-4o with questions such as: "Is \textit{subordinate word} a typical kind of \textit{superordinate word}?", with the instruction to "give a percentage" response. The top-likelihood answer was considered the typicality percentage value. Specificity refers to the level of detail at which a concept is represented within the semantic system \citep{Lamarra2025}. To model specificity, GPT-4o was used to define the semantics for the superordinate-subordinate word pairs by asking "Is \textit{subordinate word} a kind of \textit{superordinate word}?".

\subsection{Map Training}
Maps were trained by presenting inputs to each map, a total of 850 inputs per epoch. Initial activation for an input vocabulary word $v$ was calculated based on softmax normalization and scaled similarity:
\begin{equation}
a_{0,i} = \mathrm{softmax}(\frac{|u_i-v|^2}{V_m T}), \label{eqn:present}
\end{equation}
where $u_i$ is the map unit $i$’s feature vector,  $V_m$ is the median Euclidean distance between pairs of vectors in the entire input set $V $, and $T$ is a temperature parameter that controls initial activation focus. Map activations were then updated for 100 time steps ($t$) through lateral excitation and inhibition over all map units: \begin{equation}
a_{t+1, i} = \alpha a_{t,i} + \beta \sum_{j}{a_{t,j}(w_{ij}^{+}-w_{ij}^{-})},\\[-0.5ex]
\label{eqn:step}
\end{equation}
where $\alpha$ controls how fast activation fades, $\beta$ controls the strength of lateral interactions, and $w_{ij}^{+-}$ are excitatory and inhibitory connections strengths between units $i$ and $j$ \citep{prabhakaran:cogsci25}. The weight update is a blend of the winner unit's neighborhood and lateral connections. The training process results in highly activated map units becoming similar to $v$ with focused activations from long-range inhibition and short-range excitation. This process differs from traditional SOMs, where only the neighborhood of the winner unit becomes more similar to the input. 

The model was trained for familiarity by sampling inputs by word frequency. English and Spanish word frequencies were log-normalized to handle the skewed distribution, $ln(1+f_V)$, where $f_V$ is the raw word frequency. Additionally, every input's isolation in the overall feature space was measured by the Euclidean distance to the fifth-nearest neighbor within the word corpus. The probability of selection $p_i$ was a linear combination of the relative frequency of an input word ($f_i$) within the training set and the word's normalized isolation ($iso_i)$:
\begin{equation}
    p_i = 0.7 * \frac{f_i}{\sum_{j=1}^{N} f_j} + 0.3 * \frac{\mathrm{iso}_i}{\sum_{j=1}^{N} \mathrm{iso}_j}
\end{equation}
Following the predicted magnification law ($f_i^{2/3}$) \citep{Ritter1986}, more frequent input regions were expected to receive proportionally more representational area. Every 10 epochs, under- ($<\frac{1}{2}$ expected units) and overrepresented ($>2\times$ expected units) inputs were identified based on minimum squared Euclidean distance from each map unit. Underrepresented words received a boosted learning rate (up to 4$\times$) proportional to their underrepresentation; overrepresented words had their farthest units retrained toward neighboring inputs, with the retrained count set by the degree of overrepresentation. $T$ was correspondingly sharpened or broadened for under- and overrepresented inputs, respectively.

All three maps had 90 by 90 units. The learning rate decreased from 0.1 to 0.03 over 500 training epochs. The standard deviation $\sigma$ of the excitatory lateral connection neighborhood decreased from 1.8 to 0.32, and the standard deviation $\sigma$ of the winner unit's neighborhood decreased from 2.25 to 0.4. Map uncertainty, measured through the likelihood of specific input $v$ at each time step $t$, was calculated as:
\begin{equation}
L(v,t) = \sum_{i \in M_V}{a_{i,t}},\label{eqn:likelihood}
\end{equation}
where $M_v$ is the set of map units $i$ that are closer to $v$ than to any other input in $V$. $L(v,t)$ gives a probability distribution for all inputs in $V$ reflecting the (negative) entropy $E_t$ for map uncertainty at time $t$ about which input $v$ is observed \citep{prabhakaran:cogsci25}. Meeting the $E_t<2.6$ produced the maximum-likelihood response $v$ at time step $t$ in the naming experiments.

\subsection{Associative Connection Training} 
The associative connections were trained at the same time as the maps. Associative connections between subordinate semantic representations and phonetic representations were trained by presenting a randomly chosen input to both maps every 10 epochs, starting at epoch 100, with a learning rate of 0.1. Map activations were continuously updated until map uncertainty changed by less than 0.05 bits between consecutive steps, as calculated in Eqn.~\ref{eqn:likelihood}, or a threshold of $t_\mathrm{max}=80$ was reached. Co-activated units formed stronger connections according to $w_{ij}' = w_{ij} + \theta a_i a_j,$ where $\theta$ is the learning rate for associative connections \citep{prabhakaran:cogsci25}. 

Specificity was modeled through the associative connections of superordinate words in the phonetic map and their subordinate category members on the semantic map. The associative connection training mechanism was modified to scale the connection strength by the relative typicality of the subordinate category words, $w_{ij}' = w_{ij} + \theta\, s\, a_i a_j$, where $s$ is the typicality scale. For a subordinate's typicality value, $s$ is the softmax weight over the category members:
\begin{equation}
s = \frac{\exp\!\big((t_{\min}-t_j)/t_T\big)}
        {\sum_{k}\exp\!\big((t_{\min}-t_k)/t_T\big)},
\qquad
t_T = \ln(N+1),
\end{equation}
where $t_j$ is the typicality of the subordinate $j$, $t_{\min}$ is the minimum typicality among the $N$ members of the category, and $s$ sums to $1$ across members. Thus, typical category words formed stronger connections than atypical words. Post-synaptic normalization was applied after each epoch to maintain balanced associative input across semantic units. 


\section{Naming Experiments}
\label{naming_exp}
HB participants were divided into bilingual groups based on their L1 and L2, as well as their age of acquisition (AoA; early or late) in L2, forming 4 groups. BiLex models were trained for each bilingual group, accounting for L1 and L2 using AoA as a phonetic map training parameter.

\subsection{Human Data}
Healthy Spanish-English bilinguals (HBs) were recruited at Boston University as part of the larger BiLex study. Participants completed a language use questionnaire (LUQ) reporting their L1/L2 status, AoA for L2, lifetime language exposure, self-rated proficiency, and other language metrics. Lexical retrieval was assessed with the BNT, in which participants are presented with line-drawn pictures of increasing difficulty and asked to spontaneously name each item \citep{Kaplan1976}. BNT errors were classified as either: no response, neologism, unrelated (e.g.\ "ball" $\rightarrow$ "cow"), circumlocution, coordinate (word substitution from the same category; e.g.\ "dog" $\rightarrow$ "cat"), superordinate (word substitution from a higher category level; e.g.\ "tool" $\rightarrow$ "wrench"), associative (response with semantic association; e.g.\ "pillow" $\rightarrow$ "bed"), mixed, or phonological ("fig" $\rightarrow$ "pig") \citep{LETHLEAN1994207} \citep{MARIEN1998447}. All errors except for circumlocutions and neologisms were modeled with BiLex. 

\subsection{Model Naming}
Bilingual naming performance was simulated across four conditions: L1-Spanish L2-English early-AoA, L1-Spanish L2-English late-AoA, L1-English L2-Spanish early-AoA, and L1-English L2-Spanish late-AoA. L2 AoA $\geq15$ years old was set as the cutoff for late-AoA. AoA was modeled by L2-map training start time (early-AoA: $L2_\mathrm{Start}=50$, late-AoA: $L2_\mathrm{Start}=250$ epochs). Sampling, learning rate, neighborhood size, and other map training parameters were consistent across L1/L2 phonetic maps. 

BNT-item word frequency in English was matched with that in the Clearpond and COCA databases. Forty out of 60 of the BNT items were present in the word corpus. The remaining 20 items were frequency-matched with map vocabulary \citep{Nicholas01091989}. Naming was simulated by presenting the BNT item as input to the semantic map. After the semantic map produced an initial map activation, activations were updated through lateral interactions, both short-range excitation and long-range inhibition. Associative connections propagated activations bidirectionally between the semantic map and both L1/L2 phonetic maps, and between Spanish and English maps as $a_{t,j} = a_{t,j} + \gamma_t \sum_{i} a_{t,i}\, w_{ij}, \
\gamma_t = \gamma_0\, e^{-t/\tau},$ where $w_{i,j}$ is the associative connection weight between map units $i$ and $j$, and $\gamma_t$ determines the strength of interactions between maps. Cross-map propagation, $\gamma_t = \gamma_0\, e^{-t/\tau}$ (initial strength $\gamma_0$, decayed over settling steps $t$ with time constant $\tau$), was strong early, enabling interactions between semantic and phonetic maps toward the target word, then faded so that the target phonetic map sharpens into a clean winner relying on its own lateral dynamics. Naming simulations were repeated in both L1 and L2 across all 4 training conditions. 

Since lateral connections make it possible to model response times through settling, different values for the maximum number of steps to produce a response, $t_\mathrm{max} =  [75, 100, 150, 200, 400, 600$, $800, 1000]$, were tested for both naming accuracy and error types produced by the model. If the entropy of the target language map fell below  $E_t<2.6$ before $t_\mathrm{max}$ was reached, the maximum-likelihood response was produced as the named item. Superordinate errors were determined from GPT-derived category members, coordinate and associative errors by Euclidean distance from response to target in semantic feature space and category membership, and phonological errors by Euclidean distance from response to target in phonetic feature space. For each response category, percentages were computed per subject and compared against the model's fixed per-category value using a one-sample Wilcoxon rank-signed test ($\alpha = .05$, two-sided) and Holm multiple-comparison corrections, with the difference test applied only in conditions meeting a minimum sample size.  

Within each L1/L2-AoA group, item-level accuracy was compared to word frequency using a mixed-effects logistic regression to produce predicted-accuracy curves for HB English and Spanish BNT performance. Model naming performance was analyzed using the same mixed-effects logistic regression framework as the human data, and the two were compared by testing the word-frequency slope within each curve and the frequency-by-source interaction, which assessed whether the model's frequency sensitivity differed from that of humans. 

\section{Model vs.\ Human Naming Results}
\label{sect:results}
Across all four bilingual groups, BNT naming data were aggregated and analyzed for frequency-accuracy effects, as well as response and error type distributions. Accuracy and error types in the model were then compared to those of humans. This section summarizes the results; the insights from them are discussed in the next section.

\subsection{Naming Errors}
Response-type distributions for all conditions are shown in Figure \ref{fig:noresponses}, with response $t_\mathrm{max}$ set to 400 settling steps (chosen to fit the human accuracy and errors as well as possible). Across L1/L2 naming tasks and bilingual groups, the model reproduced human behavior well.

\begin{figure}
    \begin{subfigure}[t]{0.48\textwidth}
        \includegraphics[width=\linewidth,trim={0pt 0pt 0pt 5ex},clip]{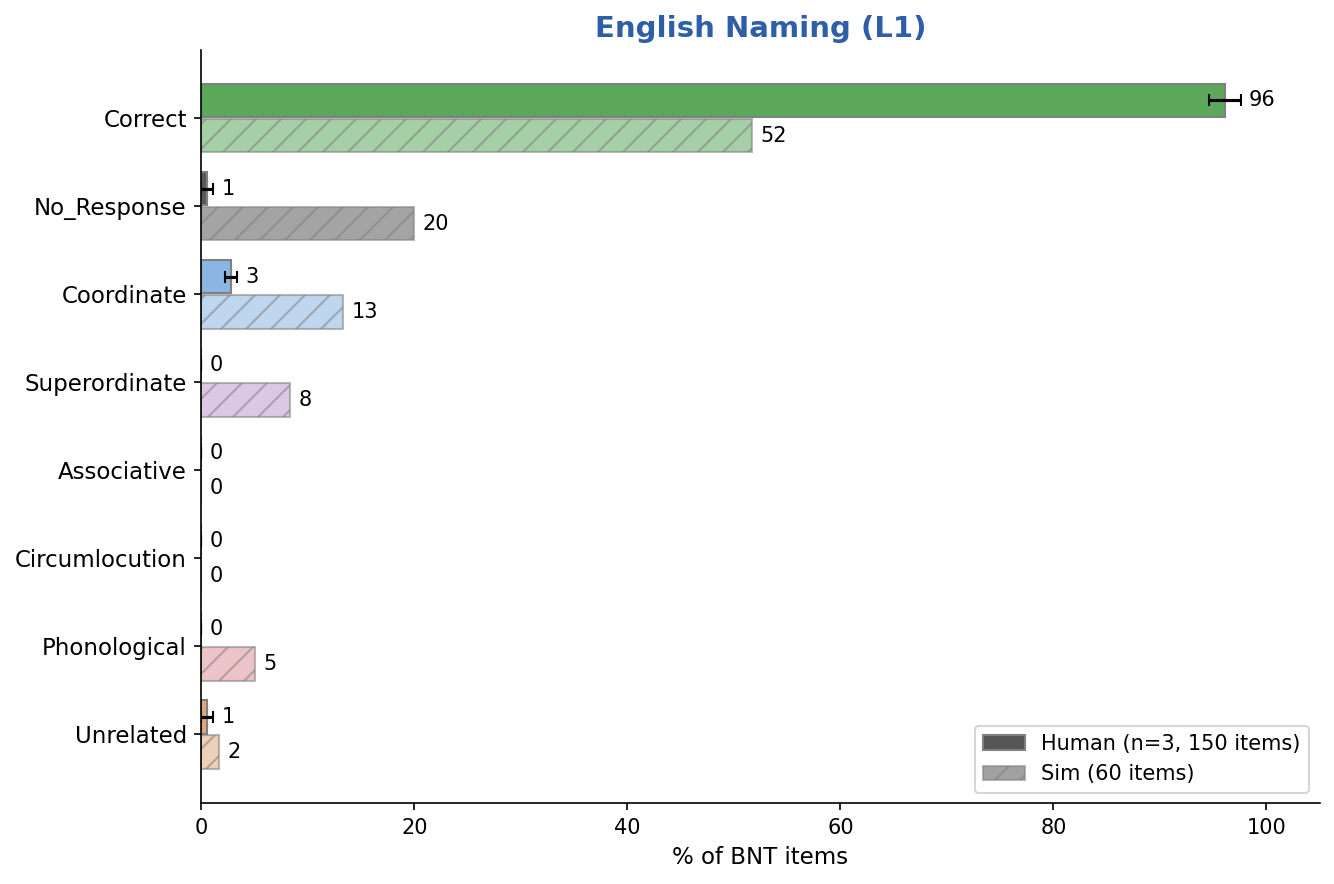}\\[-3ex]
        \caption{L2-Spanish, early-AoA: English naming}
        \label{fig:eng-early-l1}
    \end{subfigure}
    \hfill
    \begin{subfigure}[t]{0.48\textwidth}
        \includegraphics[width=\linewidth,trim={0pt 0pt 0pt 5ex},clip]{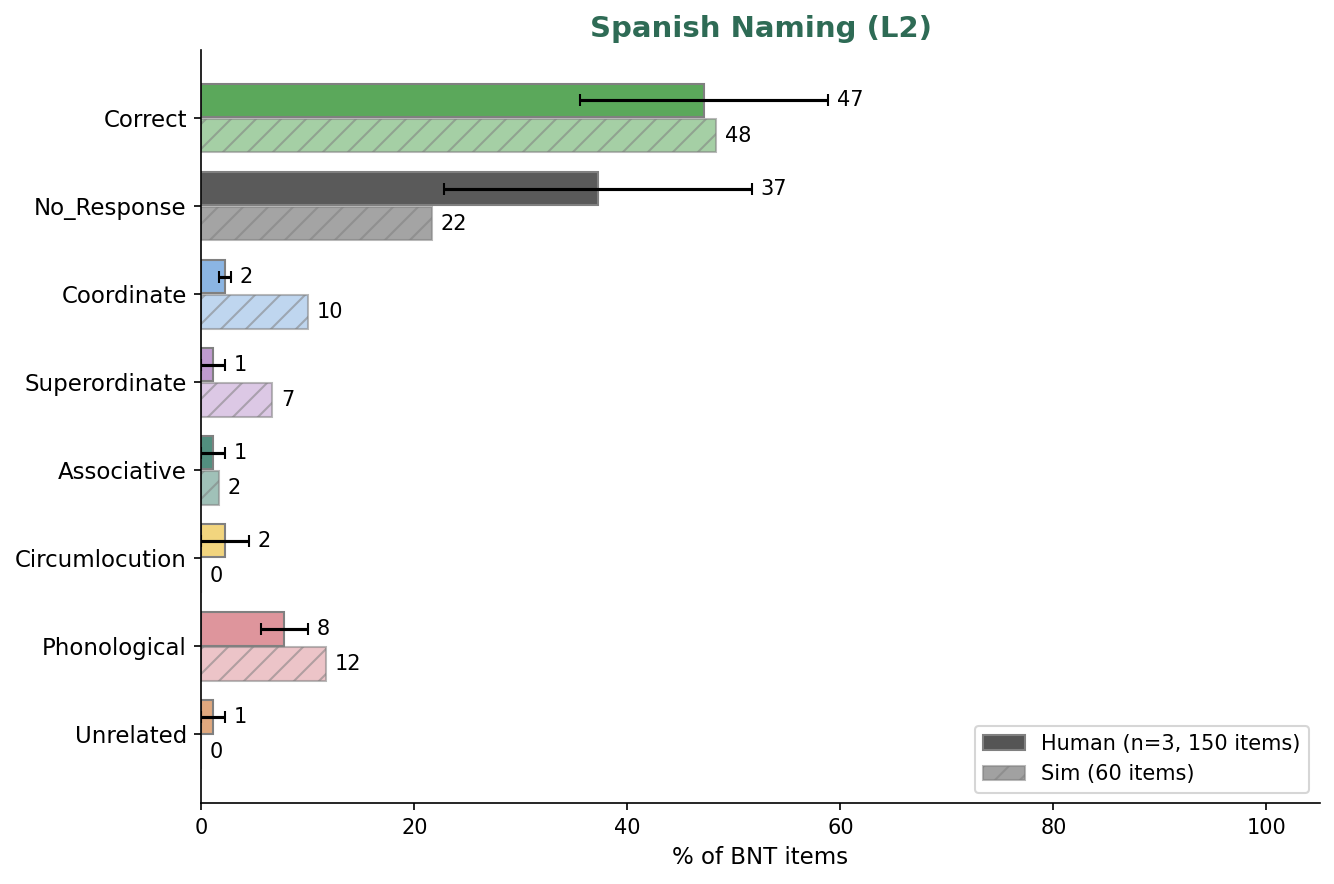}\\[-3ex]
        \caption{L2-Spanish, early-AoA: Spanish naming}
        \label{fig:eng-early-l2}
    \end{subfigure}\\[2ex]
    \begin{subfigure}[t]{0.48\textwidth}
        \includegraphics[width=\linewidth,trim={0pt 0pt 0pt 5ex},clip]{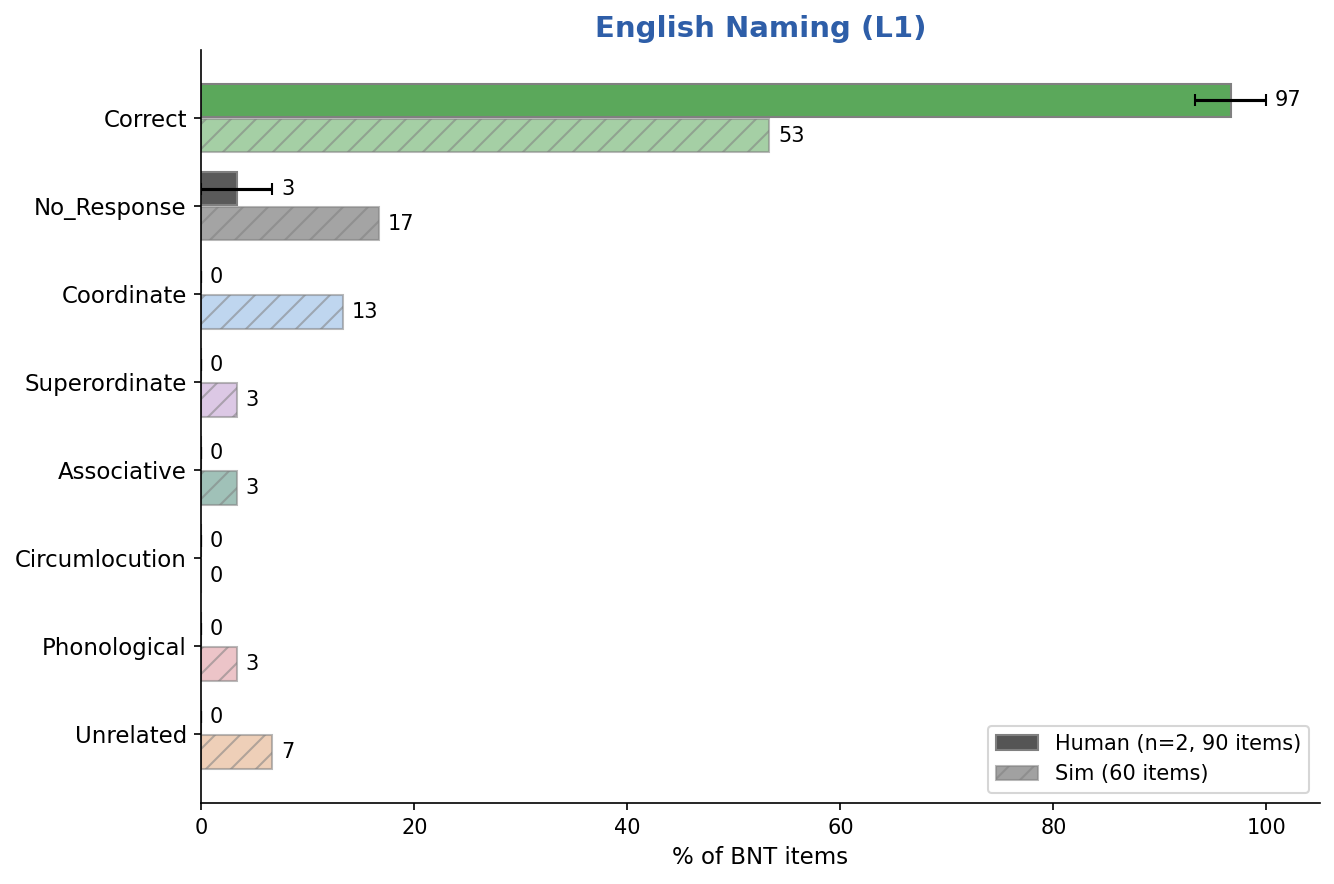}\\[-3ex]
        \caption{L2-Spanish, late AoA: English naming}
        \label{fig:eng-late-l1}
    \end{subfigure}
       \hfill
    \begin{subfigure}[t]{0.48\textwidth}
        \includegraphics[width=\linewidth,trim={0pt 0pt 0pt 5ex},clip]{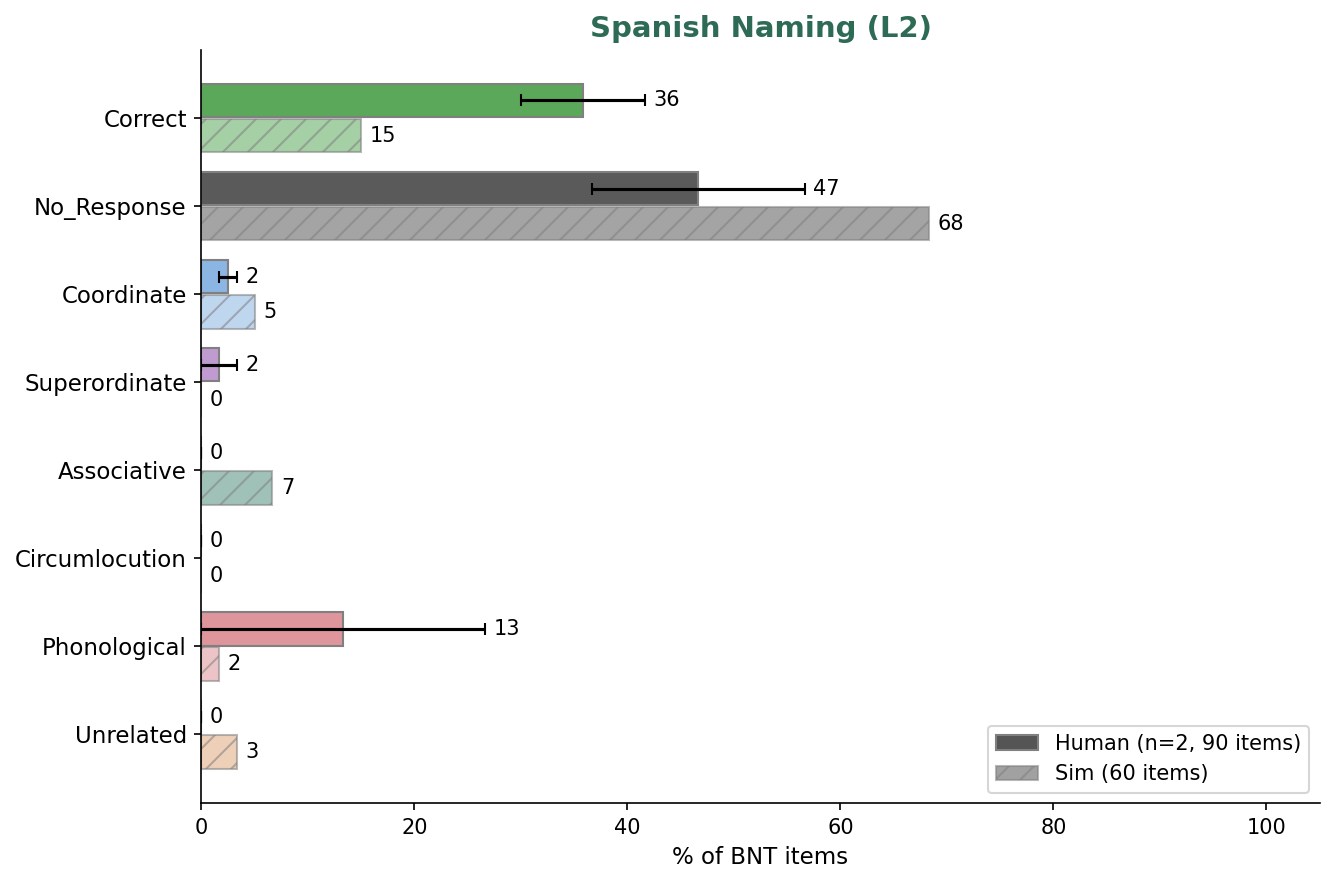}\\[-3ex]
        \caption{L2-Spanish, late AoA: Spanish naming}
        \label{fig:eng-late-l2}
    \end{subfigure}\\[2ex]
    \begin{subfigure}[t]{0.48\textwidth}
        \includegraphics[width=\linewidth,trim={0pt 0pt 0pt 5ex},clip]{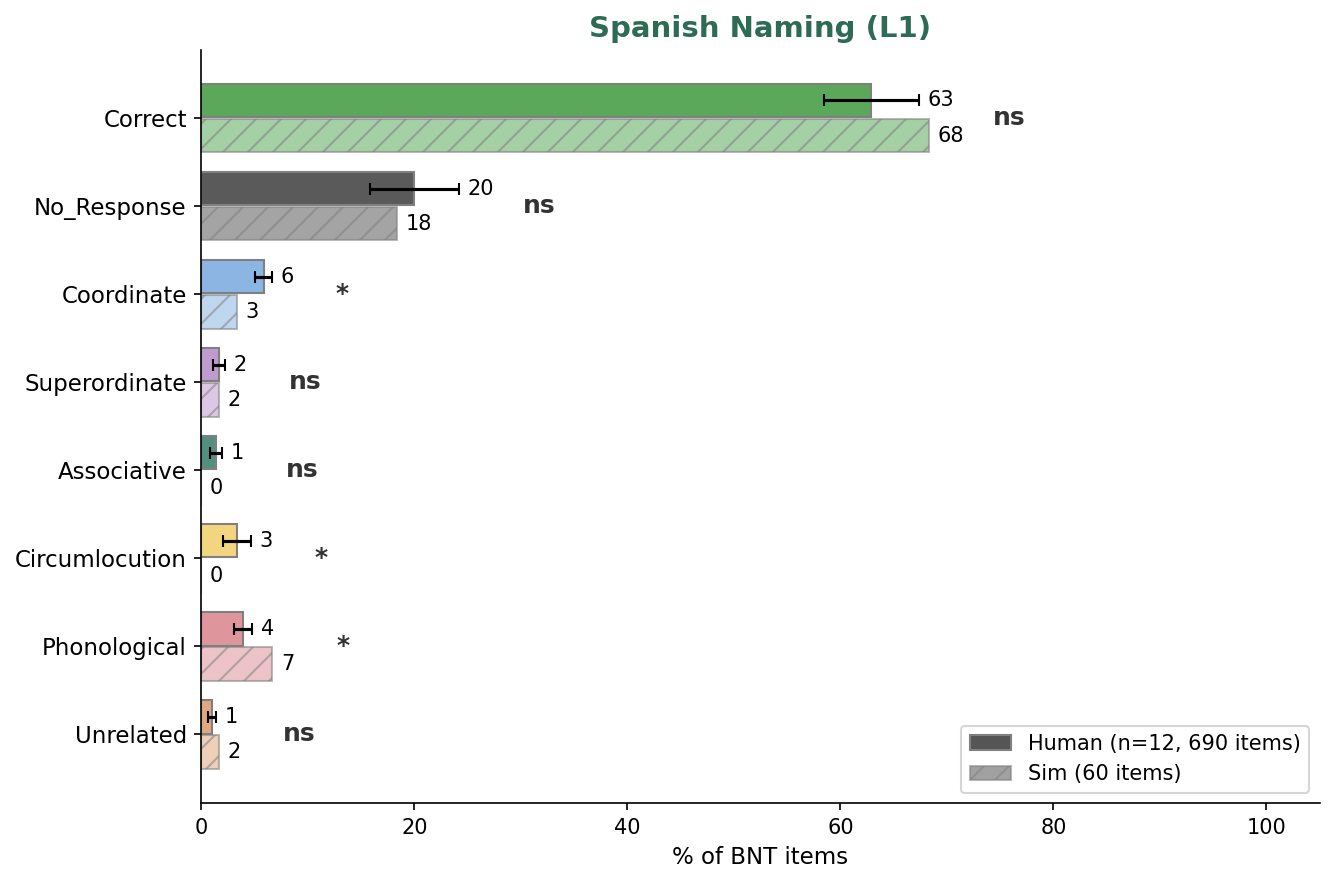}\\[-3ex]
        \caption{L2-English, early AoA: Spanish naming}
        \label{fig:spa-early-l1}
    \end{subfigure}
    \hfill
    \begin{subfigure}[t]{0.48\textwidth}
        \includegraphics[width=\linewidth,trim={0pt 0pt 0pt 5ex},clip]{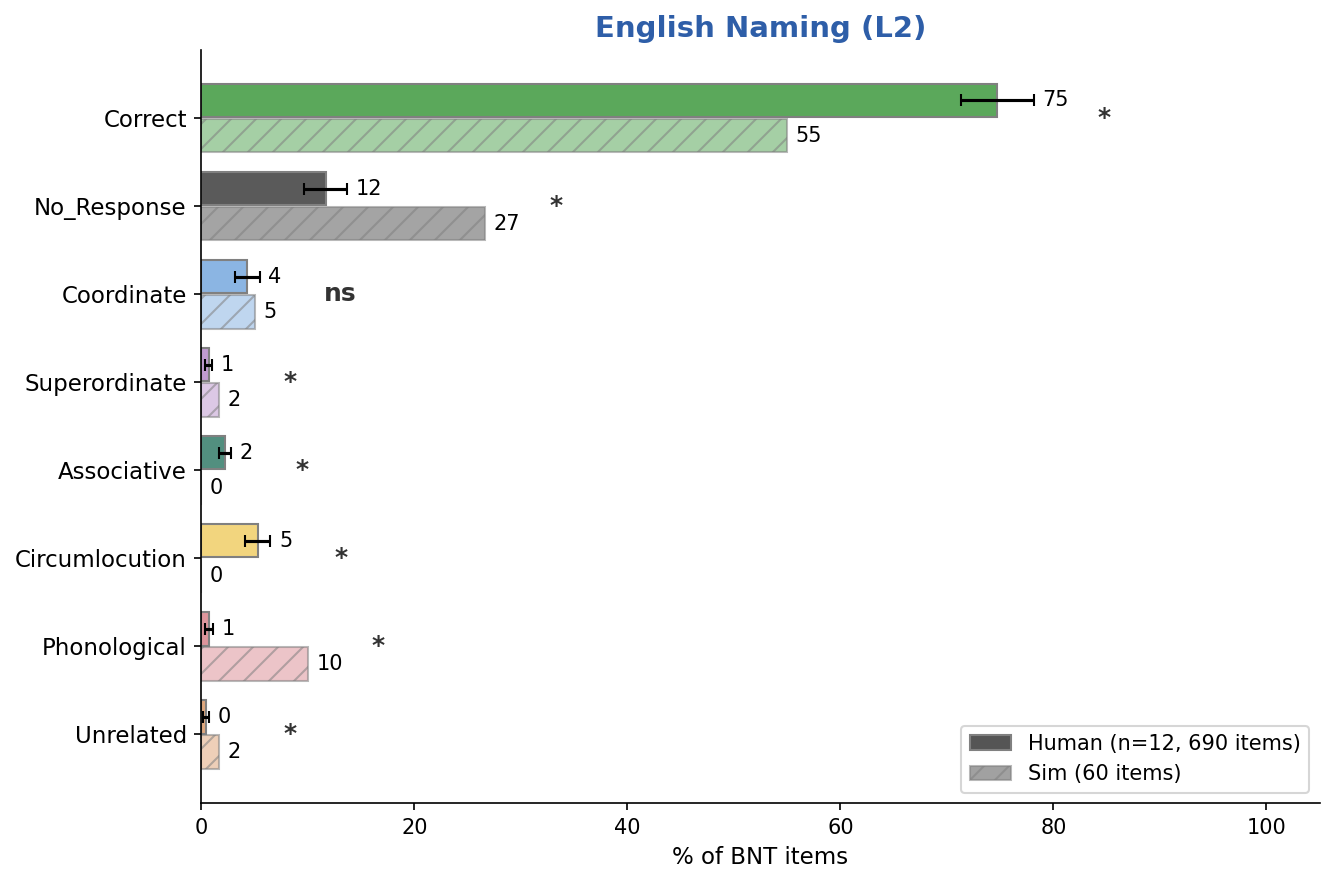}\\[-3ex]
        \caption{L2-English, early AoA: English naming}
        \label{fig:spa-early-l2}
    \end{subfigure}\\[2ex]
    \begin{subfigure}[t]{0.48\textwidth}
        \includegraphics[width=\linewidth,trim={0pt 0pt 0pt 5ex},clip]{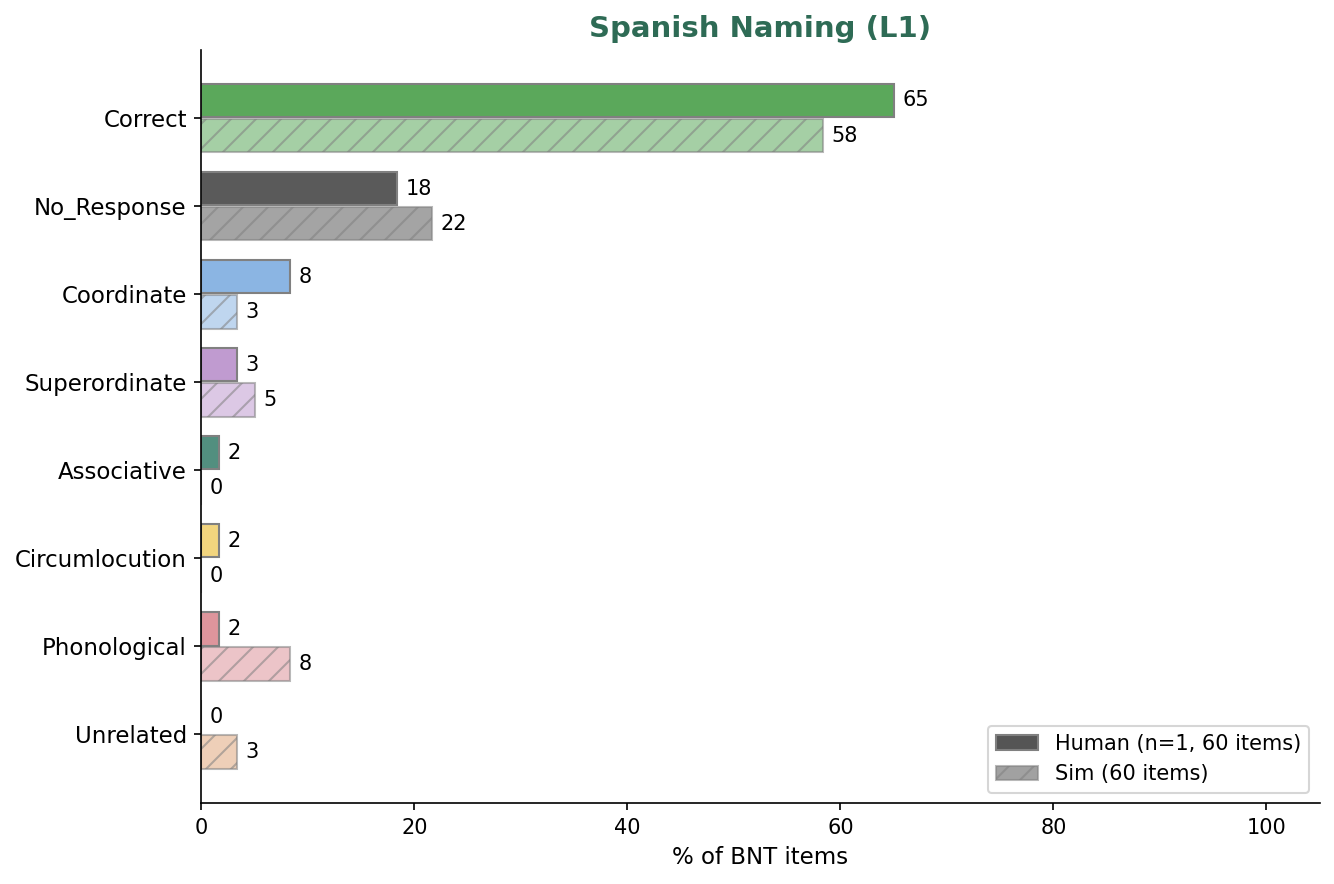}\\[-3ex]
        \caption{L2-English, late AoA: Spanish naming}
        \label{fig:spa-late-l1}
    \end{subfigure}
    \hfill
    \begin{subfigure}[t]{0.48\textwidth}
        \includegraphics[width=\linewidth,trim={0pt 0pt 0pt 5ex},clip]{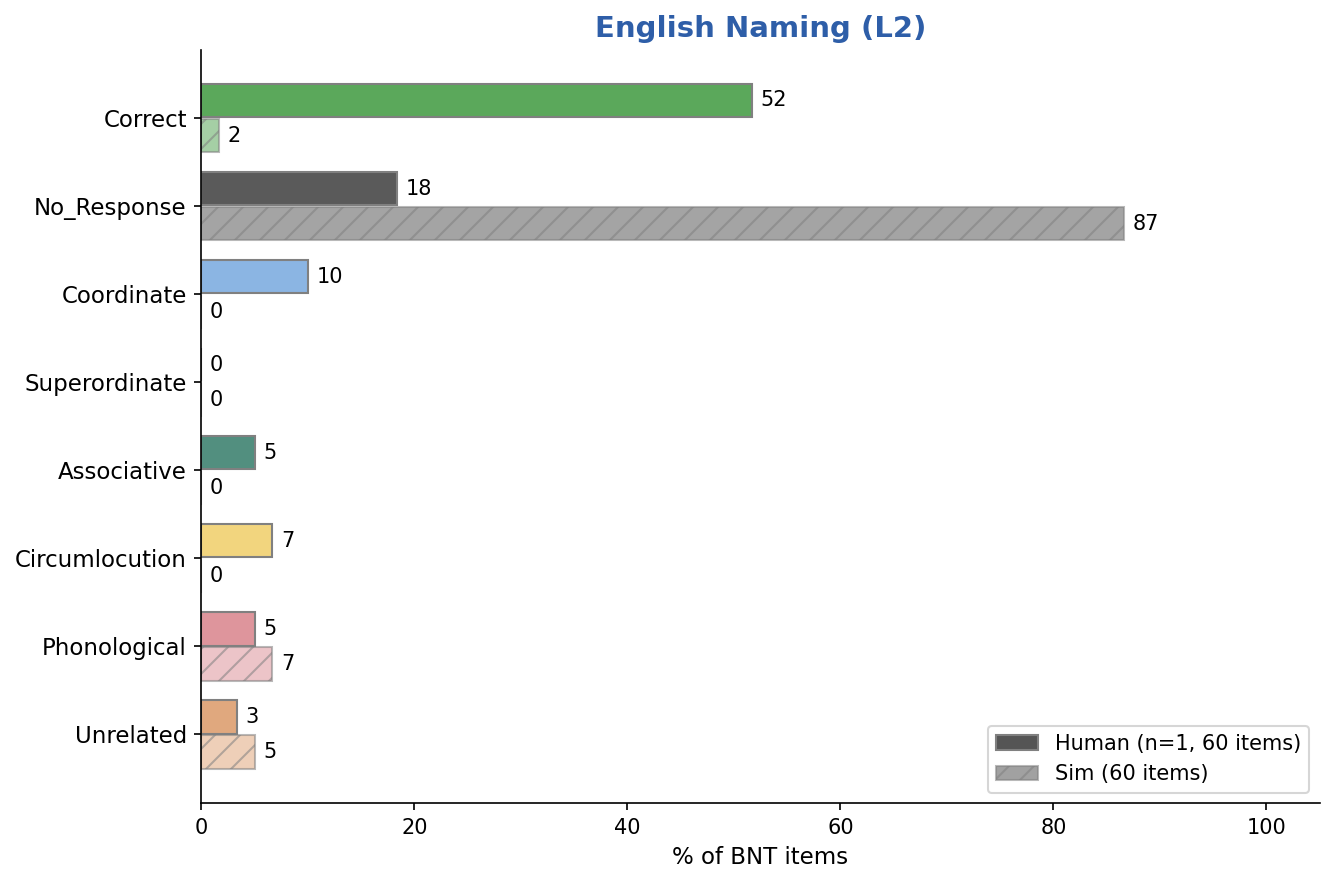}\\[-3ex]
        \caption{L2-English, late AoA: English naming}
        \label{fig:spa-late-l2}
    \end{subfigure}
    \caption {Distributions of responses in the BNT naming task in humans vs.\ the model. One-sample Wilcoxon tests were conducted on the L2-English early-AoA group (\ref{fig:spa-early-l1}, \ref{fig:spa-early-l2}), with "*" indicating statistically significant and "ns" non-significant differences. The model approximated human accuracy and error distributions well qualitatively, and also quantitatively when sufficient data were available.}
    \label{fig:bnt_errors_subfigures}
\end{figure}

With only a few subjects, it is difficult to demonstrate similarity through statistical testing. Such testing was possible only for the L2-English-early group, with $n = 12$. Some statistical similarities were indeed verified  (with Holm correction and CI of 95\%).  In Spanish naming (L1; Figure~\ref{fig:spa-early-l1}), model accuracy did not differ significantly from humans (M = 68.3\% vs.\ 62.9\%; W = 29.5, p = 1.00, r = -0.22, n.s.), nor did no-response, superordinate, associative, coordinate, or unrelated errors (all $p_{\mathrm{Holm}} > 0.11$, n.s.). However, the model produced significantly more phonological errors (6.7\% vs.\ 3.5\%; p = 0.04, r = -0.75, W = 4.0). In English naming (L2; Figure~\ref{fig:spa-early-l2}), accuracy did differ significantly (55.0\% vs.\ 74.7\%; W = 2.0, Z = 2.91, p = 0.01, r = 0.84), as did no-responses ($p_{\mathrm{Holm}} = 0.01$), associative ($p_{\mathrm{Holm}} = 0.02$), phonological ($p_{\mathrm{Holm}} = 0.004$), and unrelated errors ($p_{\mathrm{Holm}} = 0.04$).  On the other hand, coordinate (5.0\% vs.\ 4.3\%; p = 0.67, r = -0.14, W = 33.0, n.s.) and superordinate errors (1.7\% vs.\ 0.7\%; p = 0.08, r = -0.67, W = 5.0, n.s.) did not differ significantly.

The three remaining L1/L2 groups were underpowered ($n = 1$--$3$) and not statistically tested.  However, across all these groups, accuracy, no-response rates, and error distributions were qualitatively similar between humans and the model (Figures~\ref{fig:eng-early-l1}--\ref{fig:eng-early-l2}, \ref{fig:eng-late-l1}-\ref{fig:eng-late-l2}, and~\ref{fig:spa-late-l1}-\ref{fig:spa-late-l2}).

\vspace*{-1em}

\subsection{Effect of Word Frequency}

\begin{figure}
    \centering
    \begin{subfigure}[t]{0.48\textwidth}
        \includegraphics[width=\linewidth,trim={0pt 0pt 0pt 5ex},clip]{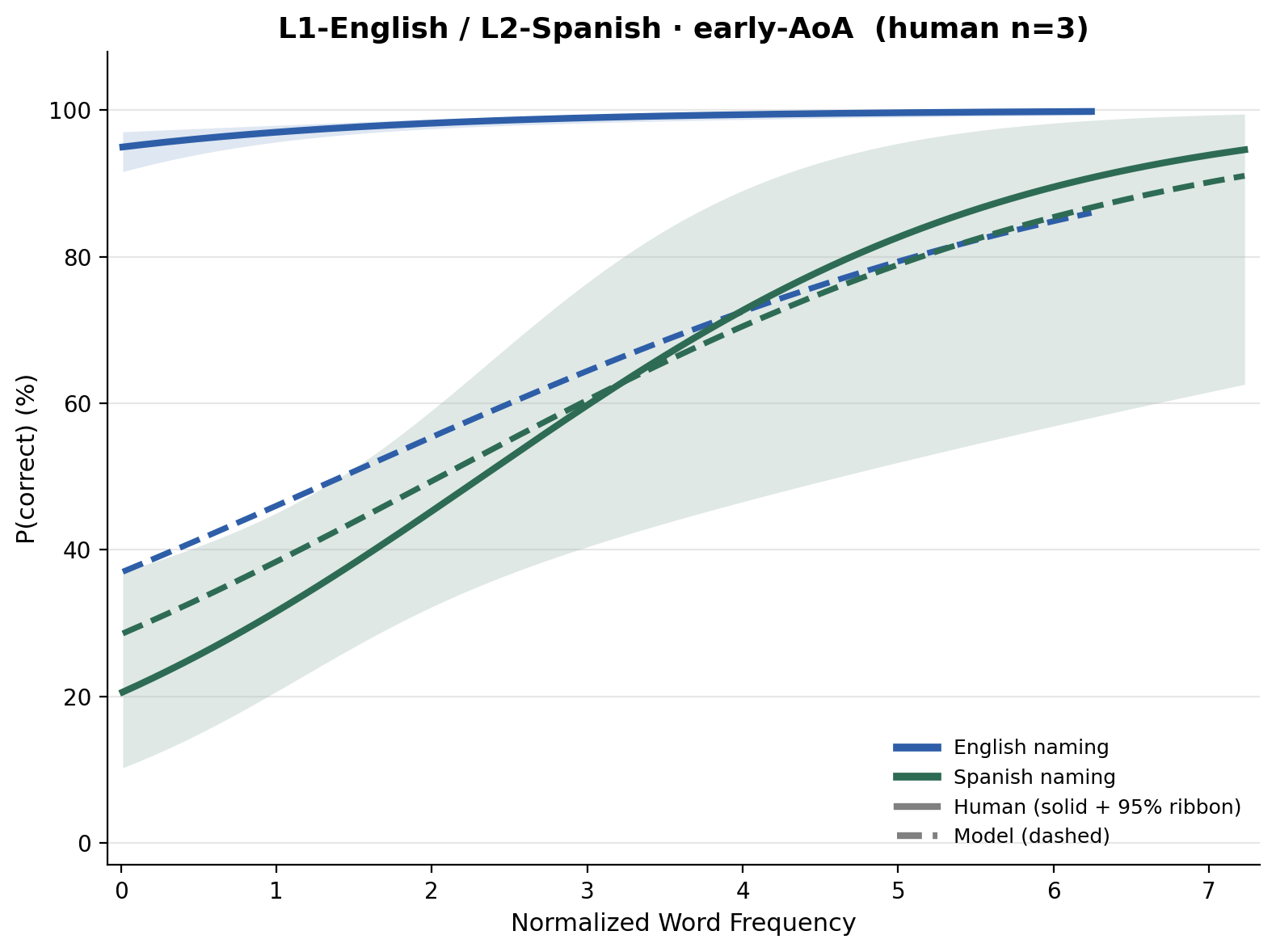}\\[-3ex]
        \caption{L2-Spanish, early AoA}
        \label{fig:spa-early-l2-freq}
    \end{subfigure}
    \hfill
    \begin{subfigure}[t]{0.48\textwidth}
        \includegraphics[width=\linewidth,trim={0pt 0pt 0pt 5ex},clip]{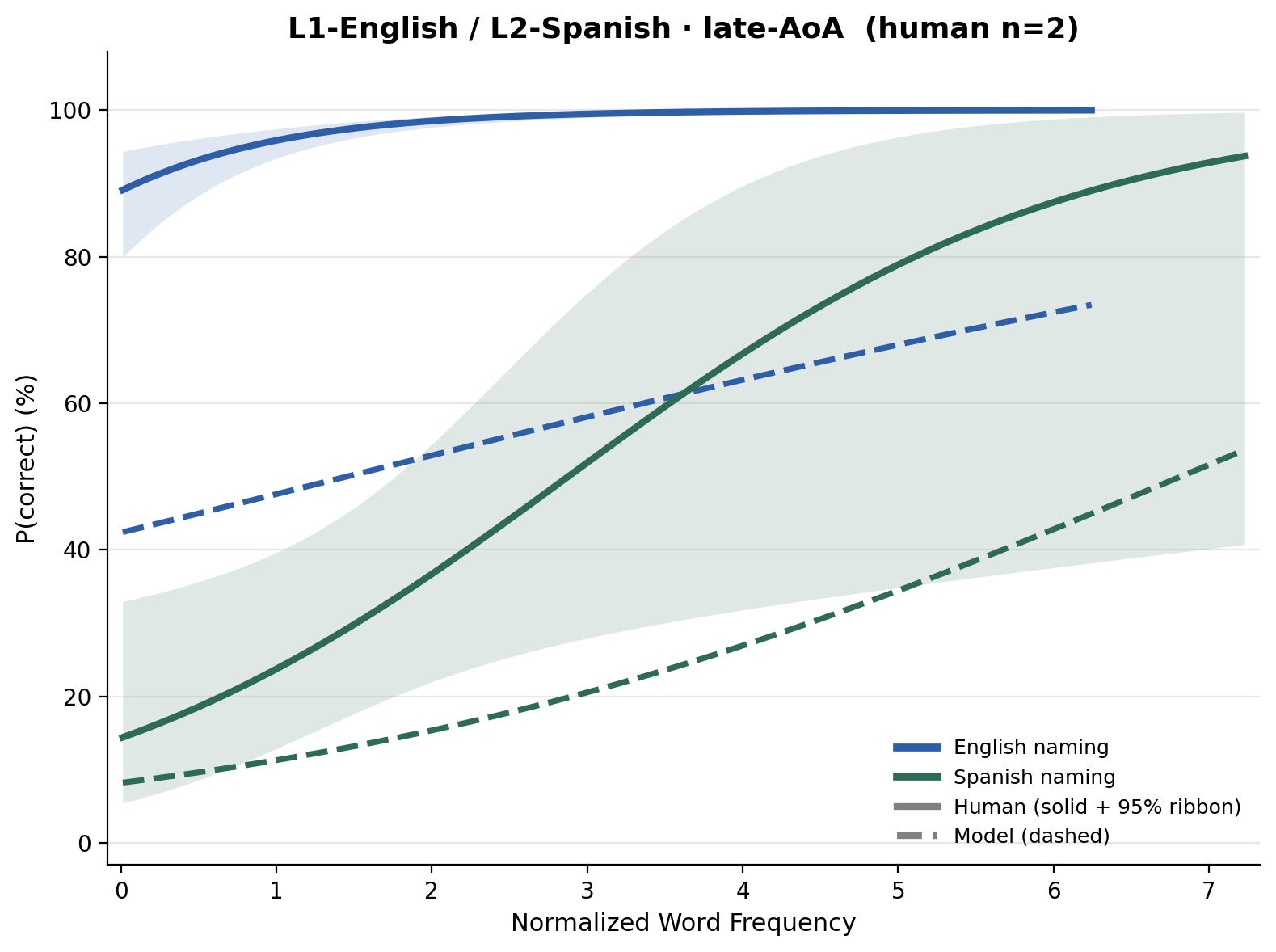}\\[-3ex]
        \caption{L2-Spanish, late AoA}
        \label{fig:spa-late-l2-freq}
    \end{subfigure}\\[2ex]
    \begin{subfigure}[t]{0.48\textwidth}
        \includegraphics[width=\linewidth,trim={0pt 0pt 0pt 5ex},clip]{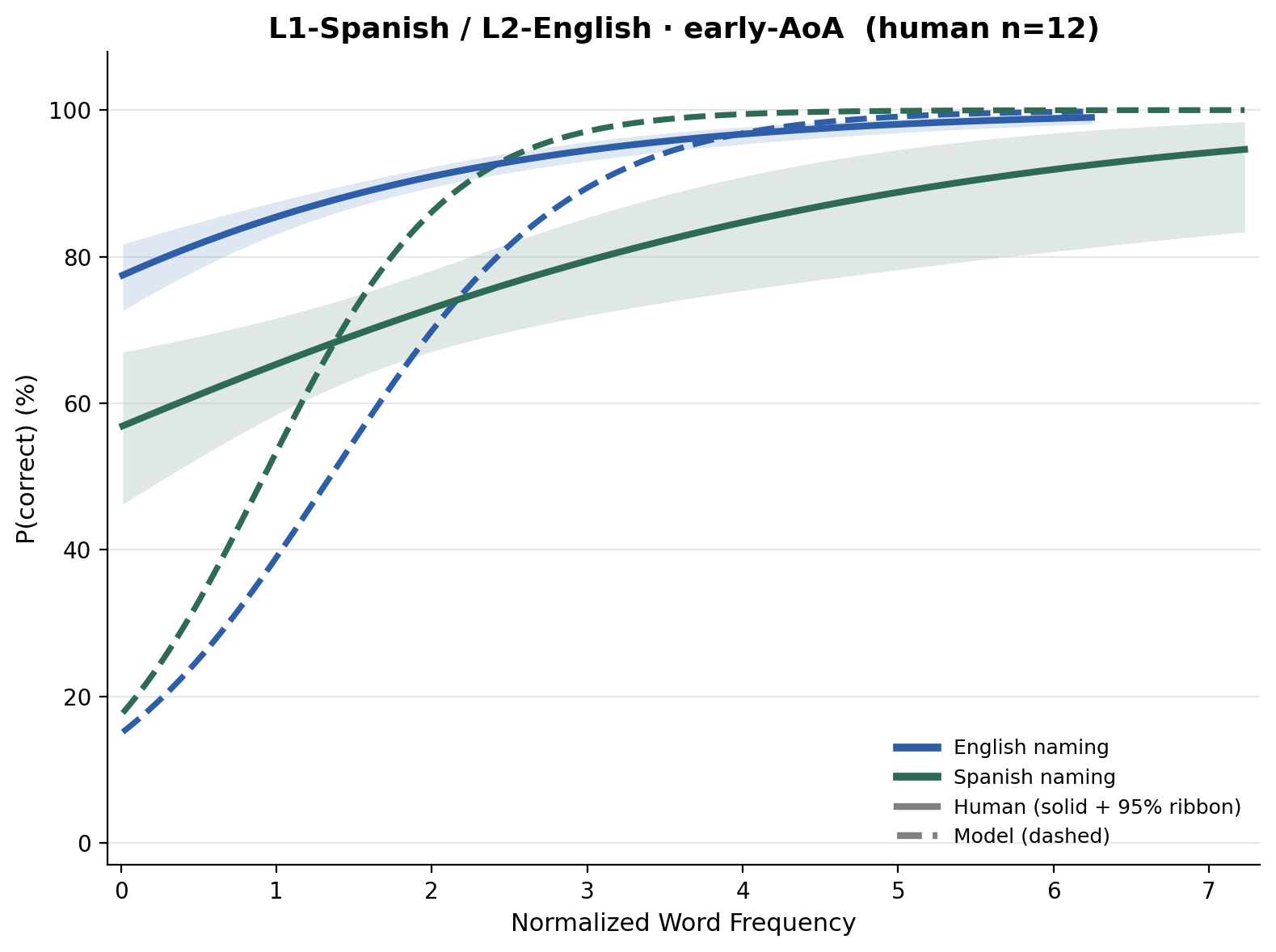}\\[-3ex]
        \caption{L2-English, early AoA}
        \label{fig:eng-early-l2-freq}
    \end{subfigure}
    \hfill
    \begin{subfigure}[t]{0.48\textwidth}
        \includegraphics[width=\linewidth,trim={0pt 0pt 0pt 5ex},clip]{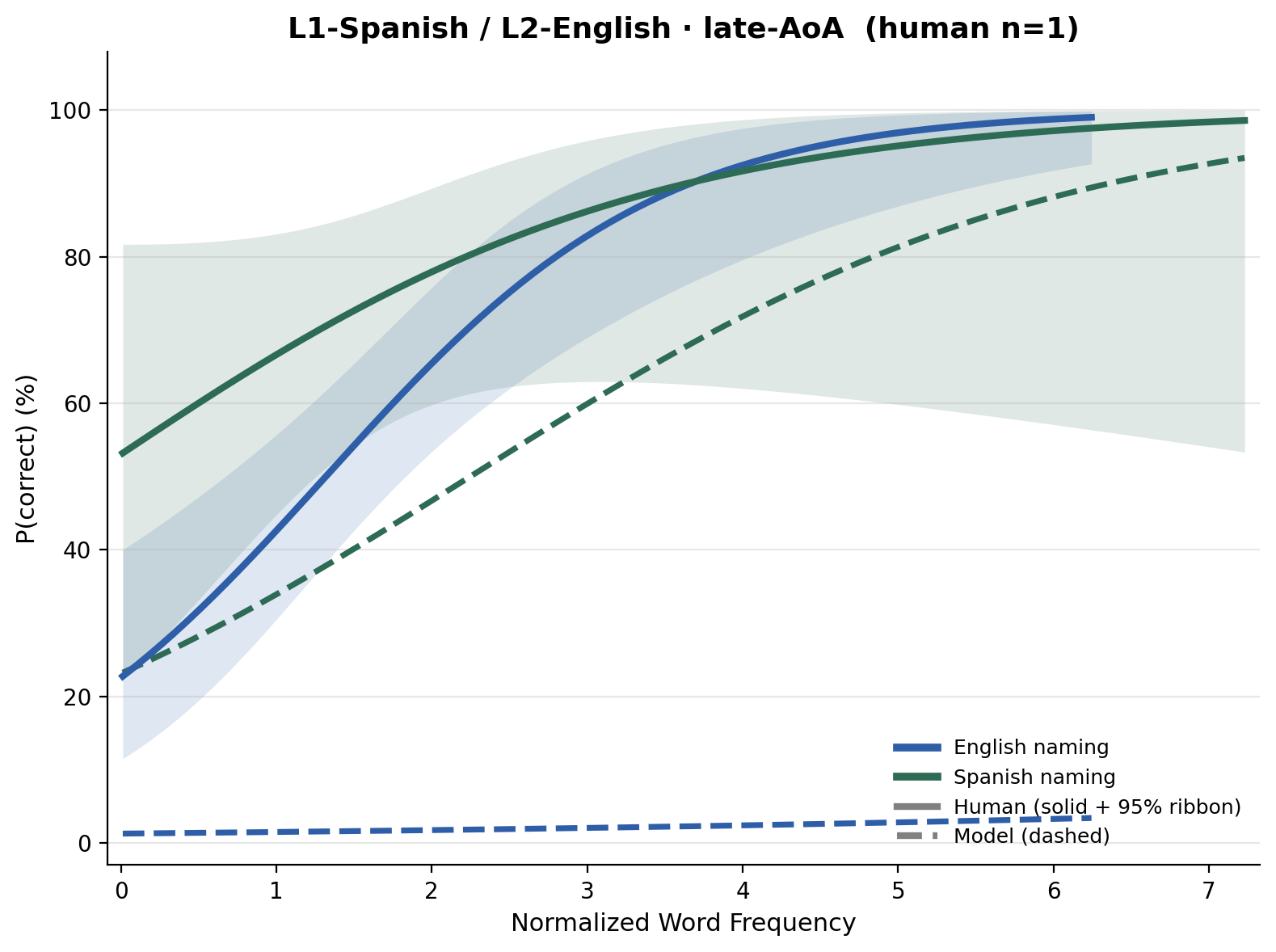}\\[-3ex]
        \caption{L2-English, late AoA}
        \label{fig:eng-late-l2-freq}
    \end{subfigure}
    \caption{Regression analysis comparing BNT word frequency to naming accuracy rates in humans vs. the model. Frequency slopes indicate L1 vs.\ L2 naming accuracy gaps. Positive word frequency vs.\ accuracy trends were present in humans and the model (significant in only L2-English-early) across bilingual groups, suggesting the model simulated human behavior of greater accuracy when naming higher-frequency words.}
    \label{fig:model_human_frequency_subpanels}
\end{figure}

In humans, frequency effects on naming were observed in all four L2 naming conditions, i.e.\ high-frequency words were named more accurately (Figure \ref{fig:model_human_frequency_subpanels}). More specifically, in L2-English-early the difference was statistically significant ($\beta = +1.14$, $[0.91, 1.37]$) and similar trend was observed for L2-Spanish-late
($\beta = +0.77$, $[0.34, 1.21]$), L2-English-late ($\beta = +1.04$, $[0.39, 1.68]$), and L2-Spanish-early ($\beta = +0.81$, $[0.46, 1.16]$). On the other hand, frequency effects in L1 were inconsistent. The L2-Spanish-late participants had the strongest effect ($\beta = +1.70$, $[0.75, 2.65]$), followed by L2-English-early ($\beta = +0.83$, $[0.64, 1.02]$). This effect was not statistically reliable for L2-Spanish-early ($\beta = +0.67$, $[-0.12, 1.46]$) and L2-English-late ($\beta = +0.55$, $[0.00, 1.10]$). A larger sample of subjects should result in a more consistent characterization of the frequency effects, but these are already sufficient as a modeling target.

Consistent with the human data, the model produced positive slopes in all cases. In the L2-English-early group, frequency effects were statistically reliable in L1 Spanish naming and L2 English naming ($\beta = +1.84, [0.62, 3.06]$, $\beta = 1.32, [0.41, 2.23]$). In the remaining six naming cases, the slopes were smaller and not
statistically reliable. The largest effect was observed in
L2-Spanish-early ($\beta = +0.49, [−0.04, 1.01]$ for English and ($\beta = +0.44, [−0.02, 0.90]$ for Spanish). L2-Spanish-late had $\beta = +0.14, [−0.27, 0.55]$ and $\beta = +0.35, [−0.04, 0.75]$; and L2-English-late had $\beta = +0.36, [−0.04, 0.75]$ and $\beta = +0.16, [−1.09, 1.41]$.

Thus, the results indicate that the model replicates human behavior well qualitatively, and also quantitatively when sufficient human data exists. The insights offered by these results are discussed in the next section.

\section{Discussion}
\label{sect:discussion}
\begin{figure}
    \centering
    \includegraphics[scale = 0.4]{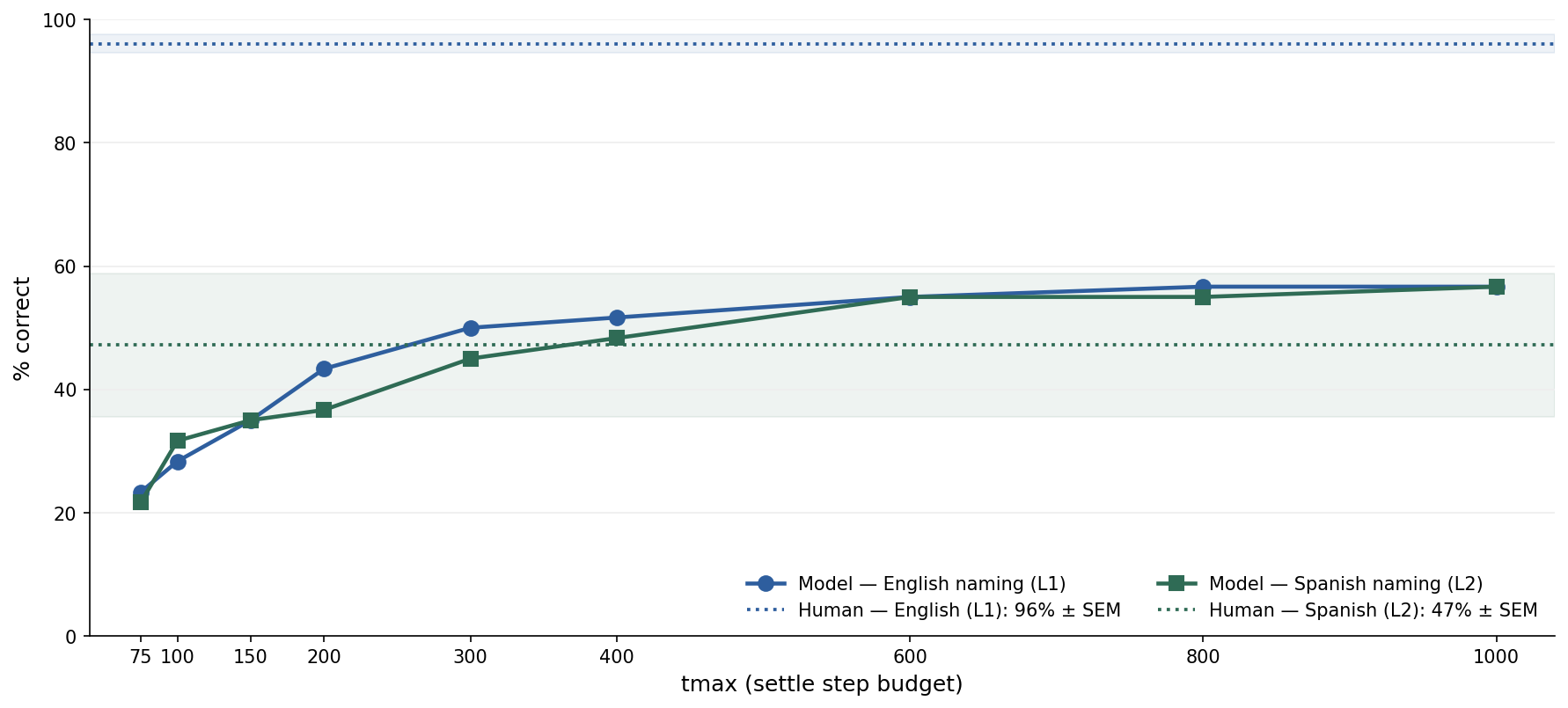}
    \caption{Adjusting settling to fit individual performance. As $t_\mathrm{max}$ increased in the L2-Spanish-early simulation, naming accuracy increased in both English and Spanish from 75 to 400 steps. At $t_\mathrm{max} = 600$ to $t_{max} = 1000$, accuracy plateaued. Thus, $t_\mathrm{max}$ can be used to fit model performance to individual subjects.}
    \label{fig:accuracyvstmax}
\end{figure}

\begin{figure}
    \centering
    \includegraphics[scale = 0.25]{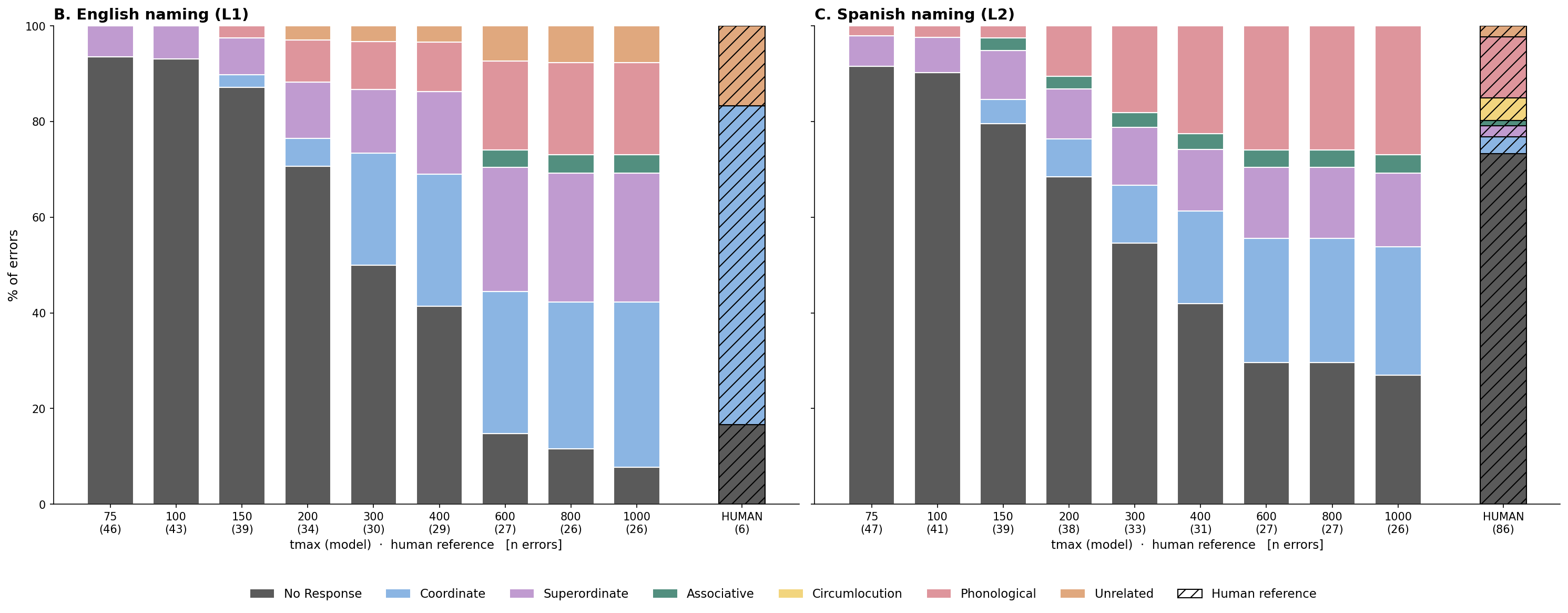}
    \caption{Distribution of naming error types in the L2-Spanish-early model. Decreasing the model's settling time ($t_\mathrm{max}$) increased the percentage of no response errors. Increasing $t_\mathrm{max}$ enabled the model to produce a wider variety of human-like errors. For all groups, $t_\mathrm{max}=400$ was chosen to balance error matching and overall naming accuracy.}
    \label{fig:noresponses}
\end{figure}

By extending BiLex with lateral connections, it was possible to model settling steps as an experimental condition where humans must perform lexical retrieval under time pressure. Adjusting $t_\mathrm{max}$ settling steps provided a parameter for improving and weakening overall naming accuracy, as shown in Figure \ref{fig:accuracyvstmax}. Increasing the maximum settling time also led to a decrease in no-responses and produced a variety of error types, as shown in Figure \ref{fig:noresponses}. It thus made it possible to fit the performance to individual subjects. Thus, the main technical contribution of this paper, i.e.\ implementing lateral connections in the BiLex model while training the maps with familiarity, typicality, and specificity, allowed the naming simulations to produce no-responses and realistic errors.  

The statistical analysis of the L2-English-early model confirmed that it can serve as a representation of bilingual lexical access. Naming accuracy did not differ significantly between the model and the human L2-English-early group in L1 (English), nor did no-response, semantic, or unrelated errors. The model also matched this group in coordinate and superordinate errors. Each error type had different underlying mechanisms: no-response errors were retrieval errors, superordinate errors from specificity-modulated associative connections, semantic errors came from the laterally-connected semantic activations, phonological errors from the lateral interactions in the phonetic space, and unrelated errors from broader naming process breakdown. Remarkably, the model reproduced these errors in both languages, suggesting that the laterally-connected model is able to capture detailed lexical effects in both dominant and non-dominant languages. 

The L2-English-early model differed from human bilingual naming in two ways. First, it produced more overall errors in L2 naming, including phonological, no-response, associative, and unrelated types. This difference is likely due to the training structure: whereas semantic and L1 maps received full training in the joint model, L2 phonetic maps began training later, resulting in less well-formed maps with noisier performance. 
Second, the model produced significantly more phonological errors than humans in both languages, i.e.\ 6.7\% in Spanish, 10.0\% in English. Phonological errors may dominate due to limited accuracy of semantic-to-phonetic associative connections, activating phonological neighbors of the target. 

Qualitative comparisons with the other groups revealed that the accuracy and error distributions were broadly similar in every case. Across groups, the relative proportions of no-response, semantic, and unrelated errors tracked closely with human performance, while phonological errors tended to be higher in the model, similar to the pattern in the statistical tests. Human participants in nearly every group and naming condition produced phonological errors from roughly 2\% to 13\%, and the models' errors were within this same range. In groups where model accuracy was weaker compared to human performance, such as L2-Spanish-late and L2-English-late English naming, no-responses were overrepresented as the primary error source. Thus, the model captured semantic-level processes in bilingual naming. Lexical retrieval was influenced by semantic map organization; in the future, with a more refined map, it should be possible to improve phonological-level processes to match human performance more accurately.

In terms of the frequency effect, the model performed consistently with the available human data across L1/L2 groups. 
A stronger frequency response could potentially be achieved with larger maps and a larger training set, giving the map more room to encode frequency in the map area.

The main limitation is the inability of the model to reach high accuracy in naming. Its accuracy is usually around 80-90\%, whereas human accuracy often reaches 96\% to 97\%. In particular, in simulations for late-AoA L2 naming, overall accuracy was extremely low (2\% and 15\%), resulting in no-responses as the primary output. Model training should be improved by scaling up the maps and the training data, resulting in better map representations for late L2. Such scaling is also important to improve the frequency effects in the model. Across all four bilingual models, word frequency-to-map unit ownership relationships were stronger in the semantic maps ($\approx r^2 = 0.6$) than in either of the phonetic maps ($\approx r^2 = 0.1-0.3$). Larger phonetic maps would allow representing frequency more accurately.
Evaluation is currently also limited and can be improved by running multiple seeds of each model to examine a distribution of responses, as well as testing multiple naming items per BNT-item frequency. Similarly, more participants across L1/L2 AoA groups are needed to make statistically significant comparisons to model naming. 

Future work will focus on individualized BiLex models; the LUQ data has metrics such as language exposure, proficiency, and confidence that can be incorporated as training parameters. Through lateral connections, response times can be modeled in human naming. Further, the BiLex model will be used to model lexical-semantic processes in bilingual semantic dementia and bilingual semantic aphasia patients. The settling process and the representation of familiarity, typicality, and specificity will allow building more accurate individualized patient models, potentially leading to more effective treatment recommendations than those in prior work.






\section{Conclusion}
By implementing lateral connections in the BiLex model and training the maps with detailed semantic effects, the model reproduced the detailed, semantically-driven error patterns characteristic of human bilingual naming. Because the model can be fit to individual performance through response time modeling and familiarity training, future work will incorporate metrics such as exposure, proficiency, and confidence to build individualized models of the bilingual lexical-semantic system. Most importantly, the BiLex model offers a framework for clinical modeling of behavior in patient populations, such as bilingual semantic dementia and aphasia, where the sources of decline and recovery can be assessed and effective treatments identified.

\subsection*{AI use statement}

In this work, we used generative AI tools for generating synthetic datasets, implementing methods, cleaning and reformatting datasets, and designing or providing feedback on research methodology or experiments. We have not used generative AI tools for the tasks of helping develop theoretical models or conceptual frameworks, assisting with translation, proposing or refining hypotheses, supporting qualitative and thematic data analysis, interpreting results, formulating mathematical claims, providing critical ingredients for proving mathematical claims, or assisting in the writing of proofs are not applicable to this work.
Additionally, we used generative AI tools for discovering research topics or identifying gaps, sourcing/searching for information, identifying relevant literature, creating or editing software code, and formatting references. We have reviewed all AI-assisted work. LLM-generated code was verified and tested for correctness, all LLM-formatted references were checked for accuracy and potential hallucinations, and a manual literature survey was conducted to check LLM-identified research topics and literature. We take responsibility for the final content of this work,
including text, claims, or artifacts produced with the aid of generative AI.


\bibliography{iclr2027_conference}
\bibliographystyle{iclr2027_conference}


\end{document}